\pdfoutput=1

\documentclass[11pt,a4paper]{article}
\usepackage{times,latexsym}
\usepackage{url}
\usepackage[T1]{fontenc}

\usepackage[acceptedWithA]{tacl2018v2}
\pdfoutput=1
\usepackage[T1]{fontenc} 
\usepackage[utf8]{inputenc} 
\usepackage{graphicx} 
\usepackage{arydshln} 
\usepackage{booktabs} 
\usepackage{amsmath}
\usepackage{amssymb}
\usepackage{color}
\usepackage{nimbusmononarrow} 
\usepackage{url}

\newcommand{\cev}[1]{\reflectbox{\ensuremath{\vec{\reflectbox{\ensuremath{#1}}}}}}
\newcommand{\example}[1]{{\ttfamily #1}}

\title{SECTOR: A Neural Model for Coherent Topic Segmentation \\and Classification}

\author{
  Sebastian Arnold \\ {\bf Rudolf Schneider} \\
  Beuth University of Applied \\
  Sciences Berlin, Germany \\
  {\sf \{sarnold, ruschneider\}}@\\{\sf beuth-hochschule.de}
  \And
  Philippe Cudré-Mauroux \\
  University of Fribourg \\
  Fribourg, Switzerland \\
  {\sf pcm@unifr.ch}
  \And
  Felix A. Gers \\ {\bf Alexander L\"oser} \\
  Beuth University of Applied \\
  Sciences Berlin, Germany \\
  {\sf \{gers, aloeser\}}@\\{\sf beuth-hochschule.de}
}

\date{}

\begin{document}
\maketitle

\begin{abstract}
When searching for information, a human reader first glances over a document, spots relevant sections and then focuses on a few sentences for resolving her intention. However, the high variance of document structure complicates to identify the salient topic of a given section at a glance. To tackle this challenge, we present SECTOR, a model to support machine reading systems by segmenting documents into coherent sections and assigning topic labels to each section. Our deep neural network architecture learns a latent topic embedding over the course of a document. This can be leveraged to classify local topics from plain text and segment a document at topic shifts. In addition, we contribute WikiSection, a publicly available dataset with 242k labeled sections in English and German from two distinct domains: diseases and cities. From our extensive evaluation of 20 architectures, we report a highest score of 71.6\% F1 for the segmentation and classification of 30 topics from the English city domain, scored by our SECTOR LSTM model with bloom filter embeddings and bidirectional segmentation. This is a significant improvement of 29.5 points F1 compared to state-of-the-art CNN classifiers with baseline segmentation.
\end{abstract}
\section{Introduction}
\label{sec:introduction}


Today's systems for natural language understanding are comprised of building blocks that extract semantic information from the text, such as named entities, relations, topics or discourse structure. In traditional natural language processing (NLP), these extractors are typically applied to bags of words or full sentences \cite{hirschberg2015advances}. Recent neural architectures build upon pre-trained word or sentence embeddings \cite{mikolov2013efficient, le2014distributed}, which focus on semantic relations that can be learned from large sets of paradigmatic examples, even from long ranges \cite{dieng2017topicrnn}.

From a human perspective, however, it is mostly the authors themselves who help best to understand a text. Especially in long documents, an author thoughtfully designs a readable structure and guides the reader through the text by arranging topics into coherent passages \cite{glavas2016unsupervised}. In many cases, this structure is not formally expressed as section headings (e.g. in news articles, reviews, discussion forums) or it is structured according to domain-specific aspects (e.g. health reports, research papers, insurance documents).


Ideally, systems for text analytics, such as topic detection and tracking (TDT) \cite{allan2002introduction}, text summarization  \cite{huang2003applying}, information retrieval (IR) \cite{dias2007topic} or question answering (QA) \cite{cohen2018wikipassageqa} could access a document representation that is aware of both \emph{topical} (i.e. latent semantic content) and \emph{structural} information (i.e. segmentation) in the text \cite{macavaney2018characterizing}.
The challenge in building such a representation is to combine these two dimensions which are strongly interwoven in the author's mind. It is therefore important to understand topic segmentation and classification as a mutual task that requires to encode both topic information and document structure coherently.


In this article, we present \textsc{Sector}\footnote{ Our source code is available under the Apache License 2.0 at \url{https://github.com/sebastianarnold/SECTOR}}, an end-to-end model which learns an embedding of latent topics from potentially ambiguous headings and can be applied to entire documents to predict local topics on sentence level. Our model encodes topical information on a vertical dimension and structural information on a horizontal dimension. We show that the resulting embedding can be leveraged in a downstream pipeline to segment a document into coherent sections and classify the sections into one of up to 30 topic categories reaching 71.6\% $F_1$ – or alternatively attach up to 2.8k topic labels with 71.1\% MAP. We further show that segmentation performance of our bidirectional LSTM architecture is comparable to specialized state-of-the-art segmentation methods on various real-world datasets.


To the best of our knowledge, the combined task of segmentation and classification has not been approached on full document level before. There exist a large number of datasets for text segmentation, but most of them do not reflect real-world topic drifts \cite{choi2000advances, sehikh2017topic}, do not include topic labels \cite{eisenstein2008bayesian, jeong2010multidocument, glavas2016unsupervised} or are heavily normalized and too small to be used for training neural networks \cite{chen2009global}. We can utilize a generic segmentation dataset derived from Wikipedia that includes headings \cite{koshorek2018text}, but there is also a need in IR and QA for supervised structural topic labels \cite{agarwal2009automatically, macavaney2018characterizing}, different languages and more specific domains, such as clinical or biomedical research \cite{tepper2012statistical, tsatsaronis2012bioasq} and news-based TDT \cite{kumaran2004text, leetaru2013gdelt}.


Therefore we introduce \textsc{WikiSection}\footnote{The dataset is available under the CC BY-SA 3.0 license at \url{https://github.com/sebastianarnold/WikiSection}}, a large novel dataset of 38k articles from the English and German Wikipedia labeled with 242k sections, original headings and normalized topic labels for up to 30 topics from two domains: \emph{diseases} and \emph{cities}. We chose these subsets to cover both clinical/biomedical aspects (e.g. symptoms, treatments, complications) and news-based topics (e.g. history, politics, economy, climate). Both article types are reasonably well-structured according to Wikipedia guidelines \cite{piccardi2018structuring}, but we show that they are also complementary: diseases is a typical scientific domain with low entropy, i.e. very narrow topics, precise language and low word ambiguity. In contrast, cities resembles a diversified domain, with high entropy, i.e. broader topics, common language and higher word ambiguity, and will be more applicable to e.g. news, risk reports or travel reviews.


We compare \textsc{Sector} to existing segmentation and classification methods based on latent dirichlet allocation (LDA), paragraph embeddings, convolutional neural networks (CNNs) and recurrent neural networks (RNNs). We show that \textsc{Sector} significantly improves these methods in a combined task by up to 29.5 points $F_1$ when applied to plain text with no given segmentation.


The rest of this article is structured as follows: We introduce related work in Section \ref{sec:related}. Next, we describe the task and dataset creation process in Section \ref{sec:task}. We formalize our model in Section \ref{sec:method}. We report results and insights from the evaluation in Section \ref{sec:evaluation}. Finally, we conclude in Section \ref{sec:summary}.

\begin{figure*}[t]
\includegraphics[width=1\textwidth]{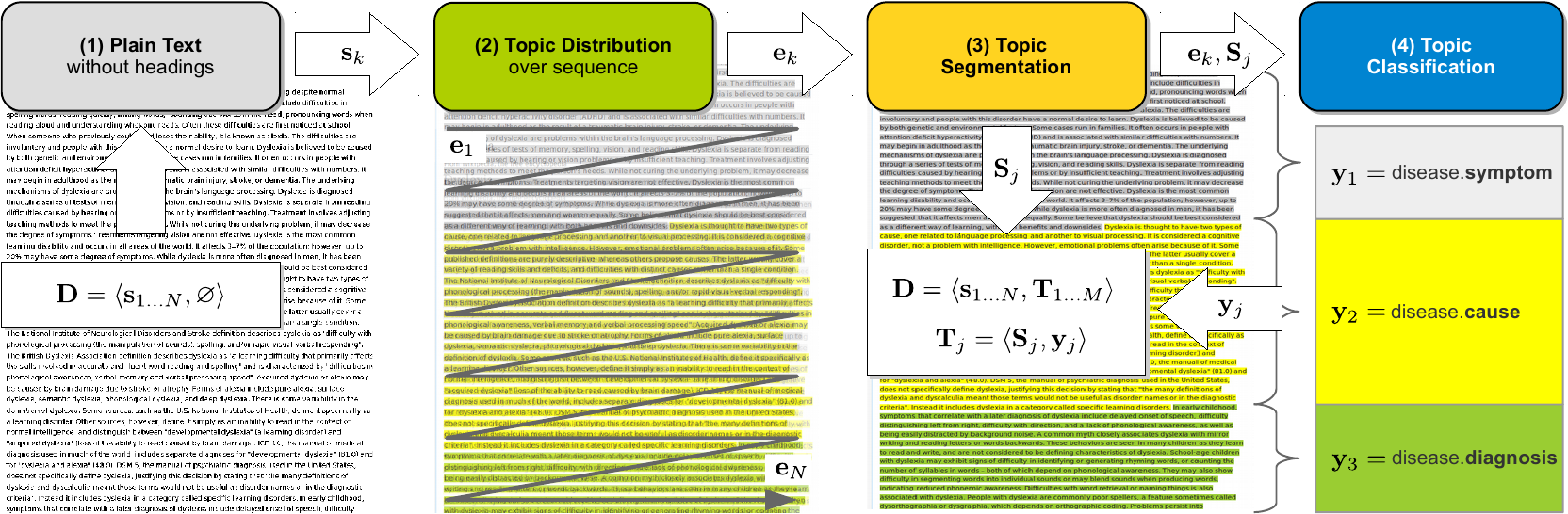}
\caption{Overview of the \textsc{WikiSection} task: (1) The input is a plain text document $\mathbf{D}$ without structure information. (2) We assume the sentences $\mathbf{s}_{1\dots N}$ contain a coherent sequence of local topics $\mathbf{e}_{1\dots N}$. (3) The task is to segment the document into coherent sections $\mathbf{S}_{1\dots M}$ and (4) to classify each section with a topic label $\mathbf{y}_{1\dots M}$.}
\label{fig:overview}
\end{figure*}
\section{Related Work}
\label{sec:related}

The analysis of emerging topics over the course of a document is related to a large number of research areas. In particular, topic modeling \cite{blei2003latent} and topic detection and tracking (TDT) 
\cite{jin1999topic} focus on representing and extracting the semantic topical content of text. Text segmentation \cite{beeferman1999statistical} is used to split documents into smaller coherent chunks. Finally, text classification \cite{joachims1998text} is often applied to detect topics on text chunks. Our method unifies those strongly interwoven tasks and is the first to evaluate the combined topic segmentation and classification task using a corresponding dataset with long structured documents.

\paragraph{Topic modeling}

is commonly applied to entire documents using probabilistic models, such as latent Dirichlet allocation (LDA) \cite{blei2003latent}. \citet{alsumait2008online} introduced an online topic model that captures emerging topics
when new documents appear.
\citet{gabrilovich2007computing} proposed the Explicit Semantic Analysis method in which concepts from Wikipedia articles are indexed and assigned to documents. Later, and to overcome the vocabulary mismatch problem, \citet{cimiano2009explicit} introduced a method for assigning latent concepts to documents. More recently, \citet{liu2016representing} represented documents with vectors of closely related domain keyphrases. 
\citet{yeh2016topic} proposed a conceptual dynamic LDA model for tracking topics in conversations.
\citet{bhatia2016automatic} utilized Wikipedia document titles to learn neural topic embeddings and assign document labels. 
\citet{dieng2017topicrnn} focused on the issue of long-range dependencies and proposed a latent topic model based on recurrent neural networks (RNNs). However, the authors did not apply the RNN to predict local topics.

\paragraph{Text segmentation} 

has been approached with a wide variety of methods. Early unsupervised methods utilized lexical overlap statistics \cite{hearst1997texttiling, choi2000advances}, dynamic programming \cite{utiyama2001statistical},
Bayesian models \cite{eisenstein2008bayesian} or 
point-wise boundary sampling \cite{du2013topic} on raw terms.

Later, supervised methods included topic models \cite{riedl2012topictiling} by calculating a coherence score using dense topic vectors obtained by LDA.
\citet{bayomi2015ontoseg} exploited ontologies to measure semantic similarity between text blocks. 
\citet{alemi2015text} and \citet{naili2017comparativec} studied how word embeddings can improve classical segmentation approaches. 
\citet{glavas2016unsupervised} utilized semantic relatedness of word embeddings by identifying cliques in a graph.

More recently, \citet{sehikh2017topic} utilized long short-term memory (LSTM) networks and showed that cohesion between bidirectional layers can be leveraged to predict topic changes. In contrast to our method, the authors focused on segmenting speech recognition transcripts on word level without explicit topic labels. The network was trained with supervised pairs of contrary examples and was mainly evaluated on artificially-segmented documents. 
Our approach extends this idea so it can be applied to dense topic embeddings which are learned from raw section headings.

\citet{wang2017learning} tackled segmentation by training a CNN to learn coherence scores for text pairs. Similar to \citet{sehikh2017topic}, the network was trained with short contrary examples and no topic objective. The authors showed that their point-wise ranking model performs well on datasets by \citet{jeong2010multidocument}.
In contrast to our method, the ranking algorithm strictly requires a given ground truth number of segments for each document 
and no topic labels are predicted.

\citet{koshorek2018text} presented a large new dataset for text segmentation based on Wikipedia that includes section headings. The authors introduced a neural architecture for segmentation which is based on sentence embeddings and four layers of bidirectional LSTM. Similar to \citet{sehikh2017topic}, the authors used a binary segmentation objective on sentence level, but trained on entire documents. Our work takes up this idea of end-to-end training and enriches the neural model with a layer of latent topic embeddings that can be utilized for topic classification.

\paragraph{Text classification} 

is mostly applied at paragraph or sentence level using machine learning methods such as Support Vector Machines \cite{joachims1998text} or, more recently, shallow and deep neural networks \cite{hoat.le2018convolutional, conneau2017very}. Notably, Paragraph Vectors \cite{le2014distributed} is an extension of word2vec for learning fixed-length distributed representations from texts of arbitrary length. The resulting model can be utilized for classification by providing paragraph labels during training.
Furthermore, \citet{kim2014convolutional} has shown that convolutional neural networks (CNNs) combined with pre-trained task-specific word embeddings achieve highest scores for various text classification tasks.

\begin{table}[t]
\centering\small
\setlength{\tabcolsep}{5.6pt}
\begin{tabular}{@{}lrrrr@{}}
\toprule
\textbf{Dataset} & \multicolumn{2}{c}{\textbf{disease}}                              & \multicolumn{2}{c}{\textbf{city}}                                 \\
language         & \multicolumn{1}{c}{\textbf{en}} & \multicolumn{1}{c}{\textbf{de}} & \multicolumn{1}{c}{\textbf{en}} & \multicolumn{1}{c}{\textbf{de}} \\ \cmidrule[0.4pt](lr{0.125em}){2-2}%
\cmidrule[0.4pt](lr{0.125em}){3-3}%
\cmidrule[0.4pt](lr{0.125em}){4-4}%
\cmidrule[0.4pt](l{0.125em}){5-5}%
total docs          & 3.6k                            & 2.3k                            & 19.5k                           & 12.5k                           \\
avg sents per doc    & 58.5                            & 45.7                            & 56.5                            & 39.9                            \\
avg sects per doc   & 7.5                             & 7.2                             & 8.3                             & 7.6                             \\
headings         & 8.5k                            & 6.1k                            & 23.0k                           & 12.2k                           \\
topics           & 27                              & 25                              & 30                              & 27                              \\
coverage         & 94.6\%                          & 89.5\%                          & 96.6\%                          & 96.1\%                          \\ \bottomrule
\end{tabular}
\caption{Dataset characteristics for \emph{disease} (German: \emph{Krankheit}) and \emph{city} (German: \emph{Stadt}).
\emph{Headings} denotes the number of distinct section and subsection headings among the documents. \emph{Topics} stands for the number of topic labels after synset clustering. \emph{Coverage} denotes the proportion of headings covered by topics; the remaining headings are labeled as \example{other}.}
\label{tab:datasets}
\end{table}

\paragraph{Combined approaches}

of topic segmentation and classification are rare to find. 
\citet{agarwal2009automatically} approached to classify sections of BioMed Central articles into four structural classes (introduction, methods, results and discussion). However, their manually-labeled dataset only contains a sample of sentences from the documents, so they evaluated sentence classification as an isolated task. 
\citet{chen2009global} introduced two Wikipedia-based datasets for segmentation, one about large cities, the second about chemical elements. While these datasets have been used to evaluate word-level and sentence-level segmentation \cite{koshorek2018text}, we are not aware of any topic classification approach on this dataset.

\citet{tepper2012statistical} approached segmentation and classification in a clinical domain as supervised sequence labeling problem. The documents were segmented using a Maximum Entropy model and then classified into 11 or 33 categories.
A similar approach by \citet{ajjour2017unit} used sequence labeling with a small number of 3–6 classes. Their model is extractive, so it does not produce a continuous segmentation  over the entire document.
Finally, \citet{piccardi2018structuring} did not approach segmentation, but recommended an ordered set of section labels based on Wikipedia articles.

Eventually, we are inspired by \emph{passage retrieval} \cite{liu2002passage} as an important downstream task for topic segmentation and classification. For example, \citet{hewlett2016wikireading} proposed WikiReading, a QA task to retrieve values from sections of long documents. The objective of TREC Complex Answer Retrieval is to retrieve a ranking of relevant passages for a given outline of hierarchical sections \cite{nanni2017benchmark}. Both tasks highly depend on a building block for local topic embeddings such as our proposed model.

\section{Task Overview and Dataset}
\label{sec:task}

We start with a definition of the \textsc{WikiSection} machine reading task shown in Figure \ref{fig:overview}. We take a document $\mathbf{D} = \langle \mathbf{S}, \mathbf{T} \rangle$ consisting of $N$ consecutive sentences $\mathbf{S} = [\mathbf{s}_1,\dots,\mathbf{s}_N]$ and empty segmentation $\mathbf{T} = \varnothing$ as input. In our example, this is the plain text of a Wikipedia article (e.g. about \example{Trichomoniasis}\footnote{\url{https://en.wikipedia.org/w/index.php?title=Trichomoniasis&oldid=814235024}}) without any section information. For each sentence $\mathbf{s}_{k}$, we assume a distribution of local topics $\mathbf{e}_k$ that gradually changes over the course of the document.

The task is to split $\mathbf{D}$ into a sequence of distinct topic sections $\mathbf{T} = [\mathbf{T}_1,\dots,\mathbf{T}_M]$, so that each predicted section $\mathbf{T}_{j} = \langle \mathbf{S}_j, \mathbf{y}_j \rangle$ contains a sequence of coherent sentences $\mathbf{S}_j \subseteq \mathbf{S}$
and a topic label $\mathbf{y}_j$ 
that describes the common topic in these sentences.
For the document \example{Trichomoniasis}, the sequence of topic labels is $\mathbf{y}_{1\dots M} = [$ \example{symptom}, \example{cause}, \example{diagnosis}, \example{prevention}, \example{treatment}, \example{complication}, \example{epidemiology} $]$.

\subsection{WikiSection Dataset}
\label{sec:dataset}

For the evaluation of this task, we created \textsc{WikiSection}, a novel dataset containing a gold standard of 38k full-text documents from English and German Wikipedia comprehensively annotated with sections and topic labels (see Table \ref{tab:datasets}). 

The documents originate from recent dumps in English\footnote{\url{https://dumps.wikimedia.org/enwiki/}\nolinkurl{20180101}} and German\footnote{\url{https://dumps.wikimedia.org/dewiki/}\nolinkurl{20180101}}. We filtered the collection using SPARQL queries against Wikidata \cite{tanon2016freebase}. We retrieved instances of Wikidata categories \emph{disease} (Q12136) and their subcategories, e.g. \example{Trichomoniasis} or \example{Pertussis}, or \emph{city} (Q515), e.g. \example{London} or \example{Madrid}.

Our dataset contains the article abstracts, plain text of the body, positions of all sections given by the Wikipedia editors with their original headings (e.g. \example{"Causes | Genetic sequence"}) and a normalized topic label (e.g. \example{disease.cause}). We randomized the order of documents and split them into 70\% training, 10\% validation, 20\% test sets.


\subsection{Preprocessing}

To obtain plain document text, we used Wikiextractor\footnote{\url{http://attardi.github.io/wikiextractor/}}, split the abstract sections and stripped all section headings and other structure tags except newline characters and lists. 

\paragraph{Vocabulary mismatch in section headings.}
Table \ref{tab:headlines} shows examples of section headings from disease articles separated into head (most common), torso (frequently used) and tail (rare). Initially, we expected articles to share congruent structure in naming and order. Instead, we observe a high variance with 8.5k distinct headings in the diseases domain and over 23k for English cities. A closer inspection reveals that Wikipedia authors utilize headings at different granularity levels, frequently copy and paste from other articles, but also introduce synonyms or hyponyms, which leads to a \emph{vocabulary mismatch problem} \cite{furnas1987vocabulary}. As a result, the distribution of headings is heavy-tailed across all articles. Roughly 1\% of headings appear more than 25 times while the vast majority (88\%) appear 1 or 2 times only.


\begin{table}[t]
\centering\small
\centering\setlength{\tabcolsep}{4.5pt}
\begin{tabular}{@{}rllcr@{}}
\toprule
\textbf{rank}                                          & \textbf{heading $h$}                                                    & \textbf{label $\mathbf{y}$}                                                & \textbf{$H$}                                            & \textbf{freq}                                          \\ \midrule
0                                                      & Diagnosis                                                             & diagnosis                                                       & 0.68                                                  & 3,854                                                  \\ \hdashline
1                                                      & Treatment                                                             & treatment                                                       & 0.69                                                  & 3,501                                                  \\ \hdashline
\begin{tabular}[c]{@{}r@{}}2\\~\end{tabular}     & \begin{tabular}[c]{@{}l@{}}Signs and \\ Symptoms\end{tabular}         & \begin{tabular}[c]{@{}l@{}}symptom\\~\end{tabular}        & \begin{tabular}[c]{@{}c@{}}0.68\\~\end{tabular} & \begin{tabular}[c]{@{}r@{}}2,452\\~\end{tabular} \\ \hdashline
\multicolumn{5}{c}{\dots} \\ \hdashline
\begin{tabular}[c]{@{}r@{}}21\\~\end{tabular}    & \begin{tabular}[c]{@{}l@{}}Differential\\ Diagnosis\end{tabular}      & \begin{tabular}[c]{@{}l@{}}diagnosis\\~\end{tabular}      & \begin{tabular}[c]{@{}c@{}}0.23\\~\end{tabular} & \begin{tabular}[c]{@{}r@{}}236\\~\end{tabular}   \\ \hdashline
22                                                     & Pathogenesis                                                          & mechanism                                                       & 0.16                                                  & 205                                                    \\ \hdashline
23                                                     & Medications                                                           & medication                                                      & 0.14                                                  & 186                                                    \\ \hdashline
\multicolumn{5}{c}{\dots} \\ \hdashline
\begin{tabular}[c]{@{}r@{}}8,494\\~\end{tabular} & \begin{tabular}[c]{@{}l@{}}Usher Syndrome\\ Type IV\end{tabular}      & \begin{tabular}[c]{@{}l@{}}classification\\~\end{tabular} & \begin{tabular}[c]{@{}c@{}}0.00\\~\end{tabular} & \begin{tabular}[c]{@{}r@{}}1\\~\end{tabular}     \\ \hdashline
\begin{tabular}[c]{@{}r@{}}8,495\\~\end{tabular} & \begin{tabular}[c]{@{}l@{}}False Melanose\\ Lesions\end{tabular}      & \begin{tabular}[c]{@{}l@{}}other\\~\end{tabular}          & \begin{tabular}[c]{@{}c@{}}0.00\\~\end{tabular} & \begin{tabular}[c]{@{}r@{}}1\\~\end{tabular}     \\ \hdashline
\begin{tabular}[c]{@{}r@{}}8,496\\~\end{tabular} & \begin{tabular}[c]{@{}l@{}}Cognitive\\ Therapy\end{tabular}           & \begin{tabular}[c]{@{}l@{}}treatment\\~\end{tabular}      & \begin{tabular}[c]{@{}c@{}}0.00\\~\end{tabular} & \begin{tabular}[c]{@{}r@{}}1\\~\end{tabular}   \\ \bottomrule
\end{tabular}

\caption{Frequency and entropy ($H$) of top-3 head and randomly selected torso and tail headings for category diseases in the English Wikipedia.}
\label{tab:headlines}
\end{table}

\subsection{Synset Clustering}
\label{sec:synsets}
In order to use Wikipedia headlines as a source for topic labels, we contribute a normalization method to reduce the high variance of headings to few representative labels based on the clustering of BabelNet synsets \cite{navigli2012babelnet}.

We create a set $\mathcal{H}$ that contains all headings in the dataset and use the BabelNet API to match\footnote{We match lemmas of main senses and compounds to synsets of type NOUN CONCEPT.} each heading $h \in \mathcal{H}$ to its corresponding synsets $S_h \subset S$. For example, \example{"Cognitive behavioral therapy"} is assigned to synset \example{bn:03387773n}. Next, we insert all matched synsets into an undirected graph $G$ with nodes $s \in S$ and edges $e$. We create edges between all synsets that match among each other with a lemma $h' \in \mathcal{H}$. 
Finally, we apply a community detection algorithm \cite{newman2006finding} on $G$ to find dense clusters of synsets. We use these clusters as normalized topics and assign the sense with most outgoing edges as representative label, in our example e.g. \example{therapy}.

From this normalization step we obtain 598 synsets which we prune using the head/tail division rule $\mathrm{count}(s) < \frac{1}{\left\vert{S}\right\vert}\sum_{s_i \in S} \mathrm{count}(s_i)$  \cite{jiang2012head}. This method covers over 94\% of all headings and yields 26 normalized labels and one \example{other} class in the English disease dataset. Table \ref{tab:datasets} shows the corresponding numbers for the other datasets. We verify our normalization process by manual inspection of 400 randomly chosen heading–label assignments by two independent judges and report an accuracy of 97.2\% with an average observed inter-annotator agreement of 96.0\%.

\section{SECTOR Model}
\label{sec:method}

\begin{figure*}[t]
\includegraphics[width=1\textwidth]{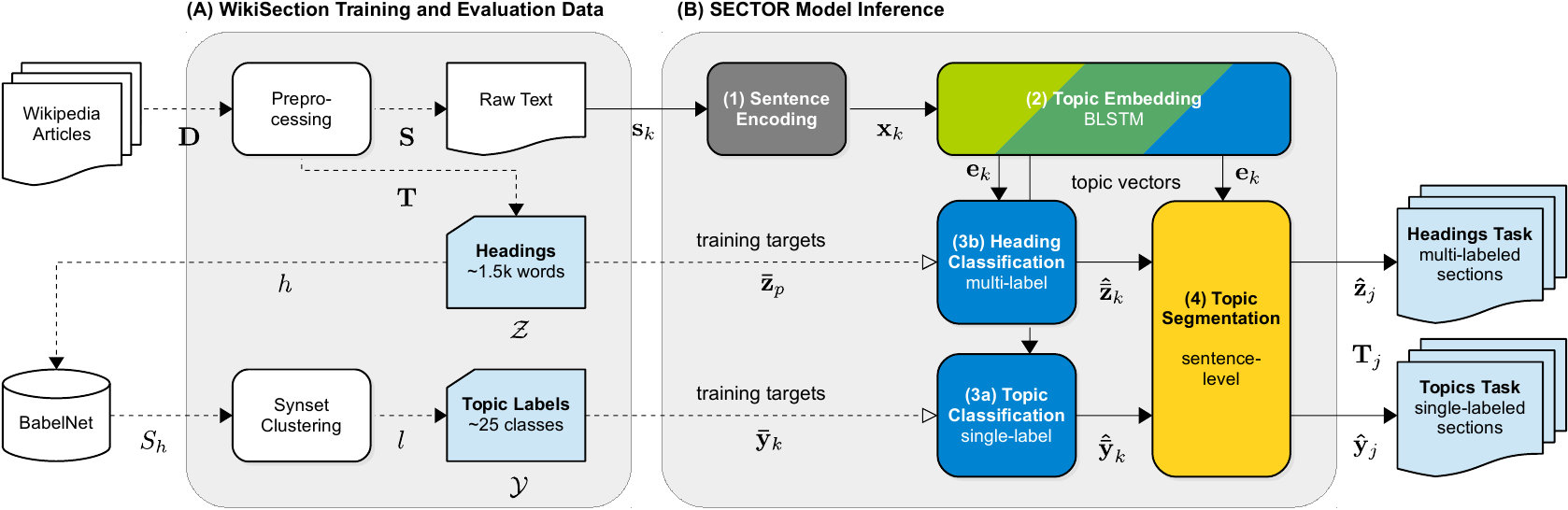}
\caption{Training and inference phase of segmentation and topic classification (\textsc{Sector}). For training (A), we preprocess Wikipedia documents to supply a ground truth for segmentation $\mathbf{T}$, headings $\mathcal{Z}$ and topic labels $\mathcal{Y}$. During inference (B), we invoke \textsc{Sector} with unseen plain text to predict topic embeddings $\mathbf{e}_k$ on sentence level. The embeddings are used to segment the document and classify headings $\mathbf{\hat{z}}_j$ and normalized topic labels $\mathbf{\hat{y}}_j$.}
\label{fig:pipeline}
\end{figure*}

We introduce \textsc{Sector}, a neural embedding model that predicts a latent topic distribution for every position in a document. 
Based on the task described in Section \ref{sec:task}, we aim to detect $M$ sections $\mathbf{T}_{0\dots M}$ in a document $\mathbf{D}$ and assign topic labels $\mathbf{y}_j = \mathrm{topic}(\mathbf{S}_j)$, where $j = 1,\dots,M$. Because we do not know the expected number of sections, we formulate the objective of our model on sentence level and later segment based on the predictions. Therefore, we assign each sentence $\mathbf{s}_k$ a sentence topic label $\mathbf{\bar{y}}_k = \mathrm{topic}(\mathbf{s}_k)$, where $k = 1,\dots,N$. Thus, we aim to predict coherent sections with respect to document context:
\begin{equation}
p(\mathbf{\bar{y}}_1, \mathinner{{\ldotp}{\ldotp}{\ldotp}}, \mathbf{\bar{y}}_N \mid \mathbf{D}) = \prod_{k=1}^{N} p(\mathbf{\bar{y}}_k \mid \mathbf{s}_1, \mathinner{{\ldotp}{\ldotp}{\ldotp}}, \mathbf{s}_N)
\end{equation}

We approach two variations of this task: for \textsc{WikiSection}-topics, we choose a single topic label $\mathbf{y}_j \in \mathcal{Y}$ out of a small number of normalized topic labels. 
However, from this simplified classification task arises an entailment problem, because topics might be hierarchically structured. For example, a section with heading \example{"Treatment | Gene Therapy"} might describe \example{genetics} as a subtopic of \example{treatment}.
Therefore, we also approach an extended task \textsc{WikiSection}-headings to capture ambiguity in a heading, 
We follow the CBOW approach \cite{mikolov2013efficient} and assign all words in the heading $\mathbf{z}_j \subset \mathcal{Z}$ as multi-label bag over the original heading vocabulary. This turns our problem into a ranked retrieval task with a large number of ambiguous labels, similar to  \newcite{prabhu2014fastxml}. It further eliminates the need for normalized topic labels.
For both tasks, we aim to maximize the log likelihood of model parameters $\Theta$
on section and sentence level:
\begin{equation}
\begin{alignedat}{2}
\mathcal{L}(\Theta) &= \sum_{j=1}^{M} \mathrm{log}\ p(\mathbf{y}_j \mid \mathbf{s}_1, \mathinner{{\ldotp}{\ldotp}{\ldotp}}, \mathbf{s}_N; \Theta) \\
\mathcal{\bar{L}}(\Theta) &= \sum_{k=1}^{N} \mathrm{log}\ p(\mathbf{\bar{y}}_k \mid \mathbf{s}_1, \mathinner{{\ldotp}{\ldotp}{\ldotp}}, \mathbf{s}_N; \Theta)
\end{alignedat}
\label{eq:objective}
\end{equation}

Our \textsc{Sector} architecture consists of four stages shown in Figure \ref{fig:pipeline}: sentence encoding, topic embedding, topic classification and topic segmentation. We now discuss each stage in more detail.

\subsection{Sentence Encoding}

The first stage of our \textsc{Sector} model transforms each sentence $\mathbf{s}_k$ from plain text into a fixed-size sentence vector $\mathbf{x}_k$ which serves as input into the neural network layers. Following \citet{hill2016learning}, word order is not critical for document-centric evaluation settings such as our \textsc{WikiSection} task. Therefore, we mainly focus on unsupervised compositional sentence representations.

\paragraph{Bag-of-words encoding.}

As a baseline, we compose sentence vectors using a weighted bag-of-words scheme. Let $\mathbb{I}(\mathbf{w}) \in \{0,1\}^{|\mathcal{V}|}$ be the indicator vector, such that $\mathbb{I}(\mathbf{w})^{(i)} = 1$ iff $\mathbf{w}$ is the $i$-th word in the fixed vocabulary $\mathcal{V}$, and let $\mathrm{tf\text{-}idf}(\mathbf{w})$ be the TF-IDF weight of $\mathbf{w}$ in the corpus. We define the sparse bag-of-words encoding $\mathbf{x}_{\mathrm{bow}} \in \mathbb{R}^{|\mathcal{V}|}$ as follows:
\begin{equation}
\mathbf{x}_{\mathrm{bow}}(\mathbf{s}) = \sum_{\mathbf{w} \in \mathbf{s}} \big( \mathrm{tf\text{-}idf}(\mathbf{w}) \cdot \mathbb{I}(\mathbf{w}) \big)
\end{equation}

\paragraph{Bloom filter embedding.}

For large $\mathcal{V}$ and long documents, input matrices grow too large to fit into GPU memory, especially with larger batch sizes. Therefore we apply a compression technique for sparse sentence vectors based on Bloom filters \cite{serra2017getting}. A Bloom filter projects every item of a set onto a bit array $\mathbb{A}(i) \in \{0,1\}^{m}$ using $k$ independent hash functions. We use the sum of bit arrays per word as compressed Bloom embedding $\mathbf{x}_{\mathrm{bloom}} \in \mathbb{N}^{m}$:
\begin{equation}
\mathbf{x}_{\mathrm{bloom}}(\mathbf{s}) = \sum_{\mathbf{w} \in \mathbf{s}} \sum_{i = 1}^{k} \mathbb{A}\big(\mathrm{hash}_i(\mathbf{w})\big)
\end{equation}

We set parameters to $m=4096$ and $k=5$ to achieve a compression factor of $0.2,$ which showed good performance in the original paper.

\paragraph{Sentence embeddings.}
We use the strategy of \citet{arora2017simple} to generate a distributional sentence representation based on pre-trained word2vec embeddings \cite{mikolov2013efficient}. This method composes a sentence vector $v_{\mathrm{emb}} \in \mathbb{R}^d$ for all sentences using a probability-weighted sum of word embeddings $v_{\mathbf{w}} \in \mathbb{R}^d$ with $\alpha = 10^{-4}$, and subtracts the first principal component $u$ of the embedding matrix $[\ v_{\mathbf{s}} : \mathbf{s} \in \mathbf{S}\ ]$:
\begin{equation}
\begin{alignedat}{2}
v_\mathbf{s} &= \frac{1}{|\mathbf{S}|} \sum_{\mathbf{w} \in \mathbf{s}} \big( \frac{\alpha}{\alpha+p(\mathbf{w})} \ v_\mathbf{w} \big) \\
\mathbf{x}_{\mathrm{emb}}(\mathbf{s}) &= v_\mathbf{s} - uu^T v_\mathbf{s}
\end{alignedat}
\end{equation}
\subsection{Topic Embedding}

\begin{figure}[t]
\includegraphics[width=1\columnwidth]{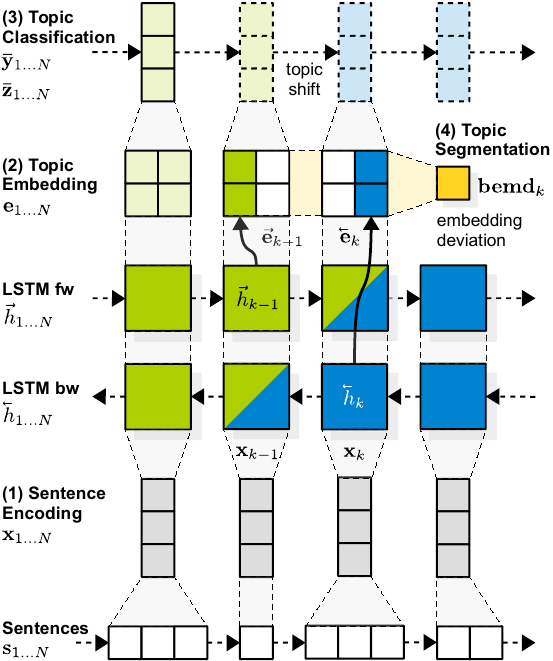}
\caption{Neural network architecture \textsc{Sector}. The recurrent model consists of stacked LSTM, embedding and output layers that are optimized on document level and later accessed during inference in stages 1–4.}
\label{fig:architecture}
\end{figure}

We model the second stage in our architecture to produce a dense distributional representation of latent topics for each sentence in the document. We use two layers of LSTM \cite{hochreiter1997long} with forget gates \cite{gers2000learning} connected to read the document in forward and backward direction \cite{graves2012supervised}. We feed the LSTM outputs to a `bottleneck' layer with tanh activation as topic embedding. Figure \ref{fig:architecture} shows these layers in context of the complete architecture.
We can see that context from left ($k-1$) and right ($k+1$) affects forward and backward layers independently. It is therefore important to separate these weights in the embedding layer to precisely capture the difference between sentences at section boundaries.
We modify our objective given in Eq. \ref{eq:objective} accordingly with long-range dependencies from forward and backward layers of the LSTM:
\begin{equation}
\begin{alignedat}{1}
\mathcal{L}(\Theta) =&
\sum_{k=1}^{N} \big( \mathrm{log}\ p(\mathbf{\bar{y}}_k \mid \mathbf{x}_{1\dots k-1}; \vec{\Theta}, \Theta' ) \\
&+ \mathrm{log}\ p(\mathbf{\bar{y}}_k \mid \mathbf{x}_{k+1\dots N}; \cev{\Theta}, \Theta' ) \big)
\end{alignedat}
\end{equation}

Note that we separate network parameters $\vec{\Theta}$ and $\cev{\Theta}$ for forward and backward directions of the LSTM, and tie the remaining parameters $\Theta'$ for the embedding and output layers. This strategy couples the optimization of both directions into the same vector space without the need for an additional loss function.
The embeddings $\mathbf{e}_{1\dots N}$ are calculated from the context-adjusted hidden states ${h'}_k$ of the LSTM cells (here simplified as $f_{\mathrm{LSTM}}$) through the bottleneck layer:
\begin{equation}
\begin{alignedat}{2}
\vec{h}_k &= f_{\mathrm{LSTM}}(\mathbf{x}_k, \vec{h'}_{k-1}, \vec{\Theta}) \\
\cev{h}_k &= f_{\mathrm{LSTM}}(\mathbf{x}_k, \cev{h'}_{k+1}, \cev{\Theta}) \\
\vec{\mathbf{e}}_k &= \mathrm{tanh}( W_{eh} \vec{h}_k + b_e )\\
\cev{\mathbf{e}}_k &= \mathrm{tanh}( W_{eh} \cev{h}_k + b_e )
\end{alignedat}
\end{equation}

Now, a simple concatenation of the embeddings $\mathbf{e}_k = \vec{\mathbf{e}}_k \oplus \cev{\mathbf{e}}_k$ can be used as topic vector by downstream applications.
\subsection{Topic Classification}

The third stage in our architecture is the output layer that decodes the class labels. To learn model parameters $\Theta$ required by the embedding, we need to optimize the full model for a training target. For the \textsc{WikiSection}-topics task, we use a simple one-hot encoding $\mathbf{\bar{y}} \in \{0,1\}^{|\mathcal{Y}|}$ of the topic labels constructed in Section \ref{sec:synsets} with a softmax activation output layer. For the \textsc{WikiSection}-headings task, 
we encode each heading as lowercase bag-of-words vector $\mathbf{\bar{z}} \in \{0,1\}^{|\mathcal{Z}|}$, such that $\mathbf{\bar{z}}^{(i)} = 1$ iff the $i$-th word in $\mathcal{Z}$ is contained in the heading, e.g. $\mathbf{\bar{z}}_k \hat{=} \{$\example{gene}, \example{therapy}, \example{treatment}$\}$. We then use a sigmoid activation function:
\begin{equation}
\begin{alignedat}{2}
\mathbf{\hat{\bar{y}}}_k &= \mathrm{softmax}( W_{ye} \vec{\mathbf{e}}_k + W_{ye} \cev{\mathbf{e}}_k + b_y )\\
\mathbf{\hat{\bar{z}}}_k &= \mathrm{sigmoid}( W_{ze} \vec{\mathbf{e}}_k + W_{ze} \cev{\mathbf{e}}_k + b_z )
\end{alignedat}
\end{equation}

\paragraph{Ranking loss for multi-label optimization.}

The multi-label objective is to maximize the likelihood of every word that appears in a heading:
\begin{equation}
\mathcal{L}(\Theta) =
\sum_{k=1}^{N} \sum_{i=1}^{|\mathcal{Z}|} \mathrm{log}\ p(\mathbf{\bar{z}}_k^{(i)} \mid \mathbf{x}_{1\dots N}; \Theta )
\end{equation} 

For training this model, we use a variation of the logistic pairwise ranking loss function proposed by \citet{santos2015classifying}. It learns to maximize the distance between positive and negative labels: 
\begin{equation}
\begin{alignedat}{2}
L &= \mathrm{log} \big( 1 + \mathrm{exp}(\gamma(m^+ - \mathrm{score}^+(\mathbf{x}))) \big) \\
&+ \mathrm{log} \big( 1 + \mathrm{exp}(\gamma(m^- +\mathrm{score}^-(\mathbf{x}))) \big)
\end{alignedat}
\label{equ:multi-class-multi-label-loss}
\end{equation} 

We calculate the positive term of the loss by taking all scores of correct labels $y^+$ into account. We average over all correct scores to avoid a too strong positive push on the energy surface of the loss function \cite{lecun2006tutorial}. For the negative term, we only take the most offending example $y^-$ among all incorrect class labels. 
\begin{equation}
\begin{alignedat}{2}
\mathrm{score}^+(\mathbf{x}) &= \dfrac{1}{|y^+|} \sum_{y \in y^+} s_\theta(\mathbf{x})^{(y)} \\
\mathrm{score}^-(\mathbf{x}) &= \underset{y \in y^-}{\operatorname{arg\,max\ }} s_\theta(\mathbf{x})^{(y)}
\end{alignedat}
\end{equation} 
Here, $s_\theta(\mathbf{x})^{(y)}$ denotes the score of label $y$ for input $\mathbf{x}$. We follow the authors and set scaling factor $\gamma=2$, margins $m^+=2.5$ and $m^-=0.5$.
\subsection{Topic Segmentation}

In the final stage, we leverage the information encoded in the topic embedding and output layers to segment the document and classify each section. 

\paragraph{Baseline segmentation methods.}
As a simple baseline method, we use prior information from the text and split sections at \emph{newline} characters (NL). Additionally, we merge two adjacent sections if they are assigned the same topic label after classification.
If there is no newline information available in the text, we use a \emph{maximum label} (max) approach: We first split sections at every sentence break, i.e. 
$\mathbf{S}_j = \mathbf{s}_k; j = k = 1,\dots, N$
and then merge all sections which share at least one label in the top-2 predictions.

\paragraph{Using deviation of topic embeddings for segmentation.}

All information required to classify each sentence in a document is contained in our dense topic embedding matrix $E = [ \mathbf{e}_{1}, \dots, \mathbf{e}_{N} ]$. We are now interested in the vector space movement of this embedding over the sequence of sentences. Therefore, we apply a number of transformations adapted from Laplacian-of-Gaussian edge detection on images \cite{ziou1998edge} to obtain the magnitude of \emph{embedding deviation} (emd) per sentence. First, we reduce the dimensionality of $E$ to $D$ dimensions using PCA, i.e. we solve $E = U \Sigma W^T$ using singular value decomposition and then project $E$ on the $D$ principal components $E_D = E W_D$. Next, we apply Gaussian smoothing to obtain a smoothed matrix $E'_D$ by convolution with a Gaussian kernel with variance $\sigma^2$. From the reduced and smoothed embedding vectors $\mathbf{e'}_{1\dots N}$ we construct a sequence of deviations $\mathbf{d}_{1\dots N}$ by calculating the stepwise difference using cosine distance:
\begin{equation}
\mathbf{d}_{k} = \mathrm{cos}(\mathbf{e'}_{k-1}, \mathbf{e'}_{k}) = \frac{\mathbf{e'}_{k-1} \cdot \mathbf{e'}_{k}}{ \parallel\mathbf{e'}_{k-1}\parallel \parallel\mathbf{e'}_{k}\parallel}
\end{equation}

\begin{figure}[b]
  \centering
  \includegraphics[width=1\columnwidth]{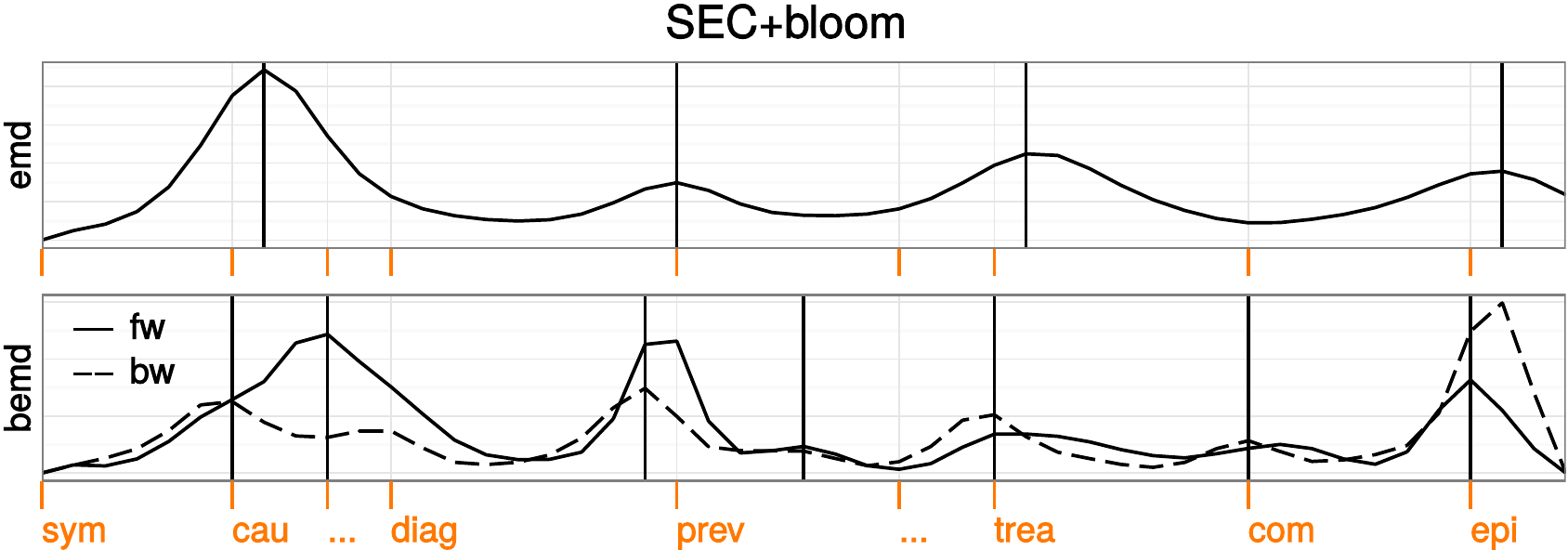}\\
  \caption{Embedding deviations $\mathbf{emd}_k$ and $\mathbf{bemd}_k$ of the smoothed \textsc{Sector} topic embeddings for example document \example{Trichomoniasis}. 
  The plot shows the first derivative of vector movement over sentences $k=1,\dots N$ from left to right.
  Predicted segmentation is shown as black lines, the axis labels indicate ground truth segmentation.
  }\label{fig:magnitudes}
\end{figure}

Finally we apply the sequence $\mathbf{d}_{1\dots N}$ with parameters $D=16$ and $\sigma = 2.5$ to locate the spots of fastest movement (see Figure \ref{fig:magnitudes}), i.e. all $k$ where $\mathbf{d}_{k-1} < \mathbf{d}_{k} > \mathbf{d}_{k+1}; k = 1\dots N$ in our discrete case. We use these positions to start a new section.

\paragraph{Improving edge detection with bidirectional layers.}
We adopt the approach of \newcite{sehikh2017topic}, who examine the difference between forward and backward layer of an LSTM for segmentation. 
However, our approach focuses on the difference of left and right topic context over time steps $k$, which allows for a sharper distinction between sections.
Here, we obtain two smoothed embeddings $\vec{\mathbf{e}'}$ and $\cev{\mathbf{e}'}$ and define the \emph{bidirectional embedding deviation} (bemd) as geometric mean of the forward and backward difference:

\begin{equation}
\mathbf{d'}_{k} = \sqrt{
	\mathrm{cos}(\vec{\mathbf{e}'}_{k-1}, \vec{\mathbf{e}'}_{k})
\cdot
	\mathrm{cos}(\cev{\mathbf{e}'}_{k}, \cev{\mathbf{e}'}_{k+1})
}
\end{equation}

After segmentation, we assign each segment the mean class distribution of all contained sentences:
\begin{equation}
\mathbf{\hat{y}}_j = \frac{1}{\mid \mathbf{S}_j \mid} \sum_{\mathbf{s}_i \in \mathbf{S}_j} \mathbf{\hat{\bar{y}}}_{i}
\end{equation}

Finally, we show in the evaluation that our \textsc{Sector} model which was optimized for sentences $\mathbf{\bar{y}}_k$ can be applied to the \textsc{WikiSection} task to predict coherently labeled sections $\mathbf{T}_{j} = \langle \mathbf{S}_j, \mathbf{\hat{y}}_j \rangle$.
\section{Evaluation}
\label{sec:evaluation}

We conduct three experiments to evaluate the segmentation and classification task introduced in Section \ref{sec:task}. The \textsc{WikiSection}-topics experiment comprises segmentation and classification of each section with a single topic label out of a small number of clean labels (25–30 topics). The \textsc{WikiSection}-headings experiment extends the classification task to multi-label per section with a larger target vocabulary (1.0k–2.8k words). This is important, because often there are no clean topic labels available for training or evaluation.
Finally, we conduct a third experiment to see how \textsc{Sector} performs across existing segmentation datasets.

\begin{table*}[t!]
\centering\small
\small \setlength{\tabcolsep}{4.2pt}
\begin{tabular}{@{}llcccccccccccc@{}}
\toprule

\multicolumn{2}{@{} l}{\begin{tabular}[c]{@{}l@{}}\textbf{WikiSection-topics}\\ single-label classification\end{tabular}} &
\multicolumn{3}{c}{\begin{tabular}[c]{@{}c@{}}\textbf{en\_disease}\\ 27 topics\end{tabular}} & 
\multicolumn{3}{c}{\begin{tabular}[c]{@{}c@{}}\textbf{de\_disease}\\ 25 topics\end{tabular}} & 
\multicolumn{3}{c}{\begin{tabular}[c]{@{}c@{}}\textbf{en\_city}\\ 30 topics\end{tabular}} & 
\multicolumn{3}{c}{\begin{tabular}[c]{@{}c@{}}\textbf{de\_city}\\ 27 topics\end{tabular}}    \\

\cmidrule[0.4pt](lr{0.125em}){3-5}%
\cmidrule[0.4pt](lr{0.125em}){6-8}%
\cmidrule[0.4pt](lr{0.125em}){9-11}%
\cmidrule[0.4pt](lr{0.125em}){12-14}%

model configuration & segm. & $P_k$ & $F_1$ & $\mathrm{MAP}$ & $P_k$ & $F_1$ & $\mathrm{MAP}$ & $P_k$ & $F_1$ & $\mathrm{MAP}$ & $P_k$ & $F_1$ & $\mathrm{MAP}$ \\ \midrule

\multicolumn{14}{@{} l}{\textbf{Classification with newline prior segmentation}} \\

PV\textgreater{}T* & NL & 35.6 & 31.7 & 47.2 & 36.0 & 29.6 & 44.5 & 22.5 & 52.9 & 63.9 & 27.2 & 42.9 & 55.5 \\
CNN\textgreater{}T* & NL & 31.5 & 40.4 & 55.6 & 31.6 & 38.1 & 53.7 & 13.2 & 66.3 & 76.1 & 13.7 & 63.4 & 75.0 \\
SEC\textgreater{}T+bow & NL & 25.8 & 54.7 & 68.4 & 25.0 & \textbf{52.7} & \textbf{66.9} & 21.0 & 43.7 & 55.3 & 20.2 & 40.5 & 52.2 \\
SEC\textgreater{}T+bloom & NL & 22.7 & \textbf{59.3} & \textbf{71.9} & 27.9 & 50.2 & 65.5 & \textbf{9.8} & \textbf{74.9} & \textbf{82.6} & 11.7 & 73.1 & 81.5 \\
SEC\textgreater{}T+emb* & NL & \textbf{22.5} & 58.7 & 71.4 & \textbf{23.6} & 50.9 & 66.8 & 10.7 & 74.1 & 82.2 & \textbf{10.7} & \textbf{74.0} & \textbf{83.0} \\ \midrule

\multicolumn{14}{@{} l}{\textbf{Classification and segmentation on plain text}} \\

C99 &  & 37.4 & n/a & n/a & 42.7 & n/a & n/a & 36.8 & n/a & n/a & 38.3 & n/a & n/a \\
TopicTiling &  & 43.4 & n/a & n/a & 45.4 & n/a & n/a & 30.5 & n/a & n/a & 41.3 & n/a & n/a \\
TextSeg &  & \textbf{24.3} & n/a & n/a & 35.7 & n/a & n/a & 19.3 & n/a & n/a & 27.5 & n/a & n/a \\
PV\textgreater{}T* & max & 43.6 & 20.4 & 36.5 & 44.3 & 19.3 & 34.6 & 31.1 & 28.1 & 43.1 & 36.4 & 20.2 & 35.5 \\
PV\textgreater{}T* & emd & 39.2 & 32.9 & 49.3 & 37.4 & 32.9 & 48.7 & 24.9 & 53.1 & 65.1 & 32.9 & 40.6 & 55.0 \\
CNN\textgreater{}T* & max & 40.1 & 26.9 & 45.0 & 40.7 & 25.2 & 43.8 & 21.9 & 42.1 & 58.7 & 21.4 & 42.1 & 59.5 \\
SEC\textgreater{}T+bow & max & 30.1 & 40.9 & 58.5 & 32.1 & 38.9 & 56.8 & 24.5 & 28.4 & 43.5 & 28.0 & 26.8 & 42.6 \\
SEC\textgreater{}T+bloom & max & 27.9 & 49.6 & 64.7 & 35.3 & 39.5 & 57.3 & \textbf{12.7} & 63.3 & 74.3 & 26.2 & 58.9 & 71.6 \\
SEC\textgreater{}T+bloom & emd & 29.7 & 52.8 & 67.5 & 35.3 & 44.8 & 61.6 & 16.4 & 65.8 & 77.3 & 26.0 & 65.5 & 76.7 \\
SEC\textgreater{}T+bloom & bemd & 26.8 & 56.6 & \textbf{70.1} & 31.7 & 47.8 & 63.7 & 14.4 & \textbf{71.6} & 80.9 & 16.8 & 70.8 & 80.1 \\
SEC\textgreater{}T+bloom+rank* & bemd & 26.8 & \textbf{56.7} & 68.8 & 33.1 & 44.0 & 58.5 & 15.7 & 71.1 & 79.1 & 18.0 & 66.8 & 76.1 \\
SEC\textgreater{}T+emb* & bemd & 26.3 & 55.8 & 69.4 & \textbf{27.5} & \textbf{48.9} & \textbf{65.1} & 15.5 & 71.6 & \textbf{81.0} & \textbf{16.2} & \textbf{71.0} & \textbf{81.1} \\ \bottomrule

\end{tabular}%
\caption{Results for topic segmentation and single-label classification on four \textsc{WikiSection} datasets. $n=$ $718$ / $464$ / $3,907$ / $2,507$ documents. 
Numbers are given as $P_k$ on sentence level, micro-averaged $F_1$ and $\mathrm{MAP}$ at segment-level. 
For methods without segmentation, we used newlines as segment boundaries (NL) and merged sections of same classes after prediction.  Models marked with * are based on pre-trained distributional embeddings.}
\label{tab:results_classes}
\end{table*}

\begin{table*}[t!]
\centering\small
\small \setlength{\tabcolsep}{3.4pt}
\begin{tabular}{@{}llcccccccccccc@{}}

\toprule
\multicolumn{2}{@{} l}{\begin{tabular}[c]{@{}l@{}}\textbf{WikiSection-headings}\\ multi-label classification\end{tabular}} & 
\multicolumn{3}{c}{\begin{tabular}[c]{@{}c@{}}\textbf{en\_disease}\\ 1.5k topics\end{tabular}} &
\multicolumn{3}{c}{\begin{tabular}[c]{@{}c@{}}\textbf{de\_disease}\\ 1.0k topics\end{tabular}} &
\multicolumn{3}{c}{\begin{tabular}[c]{@{}c@{}}\textbf{en\_city}\\ 2.8k topics\end{tabular}} &
\multicolumn{3}{c}{\begin{tabular}[c]{@{}c@{}}\textbf{de\_city}\\ 1.1k topics\end{tabular}} \\

\cmidrule[0.4pt](lr{0.125em}){3-5}%
\cmidrule[0.4pt](lr{0.125em}){6-8}%
\cmidrule[0.4pt](lr{0.125em}){9-11}%
\cmidrule[0.4pt](lr{0.125em}){12-14}%

model configuration & segm. & $P_k$ & $\mathrm{P@}1$ & $\mathrm{MAP}$ & $P_k$ & $\mathrm{P@}1$ & $\mathrm{MAP}$ & $P_k$ & $\mathrm{P@}1$ & $\mathrm{MAP}$ & $P_k$ & $\mathrm{P@}1$ & $\mathrm{MAP}$ \\ \midrule

CNN\textgreater{}H* & max & 40.9 & 36.7 & 31.5 & 41.3 & 14.1 & 21.1 & 36.9 & 43.3 & 46.7 & 42.2 & 40.9 & 46.5 \\
SEC\textgreater{}H+bloom & bemd & 35.4 & 35.8 & 38.2 & 36.9 & 31.7 & 37.8 & 20.0 & 65.2 & 62.0 & 23.4 & 49.8 & 53.4 \\
SEC\textgreater{}H+bloom+rank & bemd & 40.2 & 47.8 & 49.0 & 42.8 & 28.4 & 33.2 & 41.9 & 66.8 & 59.0 & 34.9 & 59.6 & 54.6 \\
SEC\textgreater{}H+emb* & bemd & 30.7 & \textbf{50.5} & \textbf{57.3} & \textbf{32.9} & 26.6 & \textbf{36.7} & 17.9 & \textbf{72.3} & \textbf{71.1} & 19.3 & 68.4 & \textbf{70.2} \\
SEC\textgreater{}H+emb+rank* & bemd & \textbf{30.5} & 47.6 & 48.9 & 42.9 & \textbf{32.0} & 36.4 & \textbf{16.1} & 65.8 & 59.0 & \textbf{18.3} & \textbf{69.2} & 58.9 \\
SEC\textgreater{}H+emb@fullwiki* & bemd & 42.4 & 9.7 & 17.9 & 42.7 & (0.0) & (0.0) & 20.3 & 59.4 & 50.4 & 38.5 & (0.0) & (0.1) \\ \bottomrule

\end{tabular}%
\caption{Results for segmentation and multi-label classification trained with raw Wikipedia headings. Here, the task is to segment the document and predict multi-word topics from a large ambiguous target vocabulary.}
\label{tab:results_headings}
\end{table*}

\paragraph{Evaluation datasets.}

For the first two experiments we use the \textsc{WikiSection} datasets introduced in Section \ref{sec:dataset}, which contain documents about diseases and cities in both English and German. The subsections are retained with full granularity. 
For the third experiment, text segmentation results are often reported on artificial datasets \cite{choi2000advances}. It was shown that this scenario is hardly applicable to topic-based segmentation \cite{koshorek2018text}, so we restrict our evaluation to real-world datasets that are publicly available.
The \emph{Wiki-727k} dataset by \citet{koshorek2018text} contains Wikipedia articles with a broad range topics and their top-level sections. However, it is too large to compare exhaustively, so we use the smaller \emph{Wiki-50} subset.
We further use \emph{Cities} and \emph{Elements} datasets introduced by \citet{chen2009global}, which also provide headings. These sets are typically used for word-level segmentation, so they don't contain any punctuation and are lowercased.
Finally, we use the \emph{Clinical Textbook} chapters introduced by \citet{eisenstein2008bayesian}, which do not supply headings.

\paragraph{Text segmentation models.}

We compare \textsc{Sector} to common text segmentation methods as baseline, \emph{C99} \cite{choi2000advances} and \emph{TopicTiling} \cite{riedl2012topictiling} and the state-of-the-art \emph{TextSeg} segmenter \cite{koshorek2018text}. In the third experiment we report numbers for \emph{BayesSeg} \cite{eisenstein2008bayesian} (configured to predict with unknown number of segments) and \emph{GraphSeg} \cite{glavas2016unsupervised}.

\paragraph{Classification models.} 

We compare \textsc{Sector} to existing models for single and multi-label sentence classification. Because we are not aware of any existing method for combined segmentation and classification, we first compare all methods using given prior segmentation from newlines in the text (\emph{NL}) and then additionally apply our own segmentation strategies for plain text input: maximum label (\emph{max}), embedding deviation (\emph{emd}) and bidirectional embedding deviation (\emph{bemd}).

For the experiments, we train a Paragraph Vectors (\emph{PV}) model \cite{le2014distributed} using all sections of the training sets. We utilize this model for single-label topic classification (depicted as PV\textgreater{}T) by assigning the given topic labels as paragraph IDs. Multi-label classification is not possible with this model. We use the paragraph embedding for our own segmentation strategies. We set the layer size to 256, window size to 7 and trained for 10 epochs using a batch size of 512 sentences and a learning rate of 0.025.
We further use an implementation of \emph{CNN} \cite{kim2014convolutional} with our pre-trained word vectors as input for single-label topics (CNN\textgreater{}T) and multi-label headings (CNN\textgreater{}H). We configured the models using the hyperparameters given in the paper and trained the model using a batch size of 256 sentences for 20 epochs with learning rate 0.01.

\begin{table*}[t]
\centering\small
\small \setlength{\tabcolsep}{4.6pt}
\begin{tabular}{@{}lccccccccc@{}}

\toprule

\textbf{Segmentation} & 

\multicolumn{2}{c}{\textbf{Wiki-50}} & 
\multicolumn{2}{c}{\textbf{Cities}} & 
\multicolumn{2}{c}{\textbf{Elements}} & 
\multicolumn{1}{c}{\textbf{Clinical}} \\

\cmidrule[0.4pt](lr{0.125em}){2-3}%
\cmidrule[0.4pt](lr{0.125em}){4-5}%
\cmidrule[0.4pt](lr{0.125em}){6-7}%
\cmidrule[0.4pt](lr{0.125em}){8-8}%

and multi-label classification & $P_k$ & $\mathrm{MAP}$ & $P_k$ & $\mathrm{MAP}$ & $P_k$ & $\mathrm{MAP}$ & $P_k$ \\ \midrule

GraphSeg & 63.6 & n/a & 40.0 & n/a & 49.1 & n/a & – \\
BayesSeg & 49.2 & n/a & 36.2 & n/a & \textbf{35.6} & n/a & 57.8 \\
TextSeg & \textbf{18.2*} & n/a & \textbf{19.7*} & n/a & 41.6 & n/a & \textbf{30.8} \\
SEC\textgreater{}H+emb@en\_disease & – & – & – & – & 43.3 & 9.5 & 36.5 \\
SEC\textgreater{}C+emb@en\_disease & – & – & – & – & 45.1 & n/a & 35.6 \\
SEC\textgreater{}H+emb@en\_city & 30.0 & 31.4 & 28.2 & \textbf{56.5} & 41.0 & 7.9 & – \\
SEC\textgreater{}C+emb@en\_city & 31.3 & n/a & 22.9 & n/a & 48.8 & n/a & – \\
SEC\textgreater{}H+emb@cities & 33.3 & 15.3 & 21.4* & 52.3* & 39.2 & 12.1 & 37.7 \\
SEC\textgreater{}H+emb@fullwiki & 28.6* & \textbf{32.6*} & 33.4 & 40.5 & 42.8 & \textbf{14.4} & 36.9 \\ \bottomrule

\end{tabular}%
\caption{Results for cross-dataset evaluation on existing datasets. Numbers marked with * are generated by models trained specifically for this dataset. A value of `n/a' indicates that a model is not applicable to this problem.}
\label{tab:results_segments}
\end{table*}

\paragraph{\textsc{Sector} configurations.}

We evaluate the various configurations of our model discussed in prior sections. SEC\textgreater{}T depicts the single-label topic classification model which uses a softmax activation output layer, SEC\textgreater{}H is the multi-label variant with a larger output and sigmoid activations. Other options are: bag-of-words sentence encoding (\emph{+bow}), bloom filter encoding (\emph{+bloom}) and sentence embeddings (\emph{+emb}); multi-class cross-entropy loss (as default) and ranking loss (\emph{+rank}).

We have chosen network hyperparameters using grid search on the en\_disease validation set and keep them fixed over all evaluation runs. For all configurations, we set BLSTM layer size to 256, topic embeddings dimension to 128. Models are trained on the complete train splits with a batch size of 16 documents (reduced to 8 for bag-of-words), 0.01 learning rate, 0.5 dropout and ADAM optimization. We used early stopping after 10 epochs without MAP improvement on the validation data sets. We pre-trained word embeddings with 256 dimensions for the specific tasks using word2vec on lowercase English and German Wikipedia documents using a window size of 7. All tests are implemented in Deeplearning4j and run on a Tesla P100 GPU with 16GB memory. Training a SEC+bloom model on en\_city takes roughly 5 hours, inference on CPU takes on average 0.36 seconds per document. In addition, we trained a  SEC\textgreater{}H@fullwiki model with raw headings from a complete English Wikipedia dump\footnote{excluding all documents contained in the test sets}, and use this model for cross-dataset evaluation.


\paragraph{Quality measures.} 

We measure text segmentation at sentence level using the \emph{probabilistic $P_k$ error score} \cite{beeferman1999statistical} which calculates the probability of a false boundary in a window of size $k$, lower numbers mean better segmentation. 
As relevant section boundaries we consider all section breaks where the topic label changes.
We set $k$ to half of the average segment length. 
We measure classification performance on section level by comparing the topic labels of all ground truth sections with predicted sections. We select the pairs by matching their positions using maximum boundary overlap. We report \emph{micro-averaged F1} score for single-label or \emph{Precision@1} for multi-label classification. Additionally, we measure \emph{Mean Average Precision} (MAP), which evaluates the average fraction of true labels ranked above a particular label \cite{tsoumakas2009mining}.

\subsection{Results}

\begin{figure*}[t]
  \centering
  \includegraphics[width=1\textwidth]{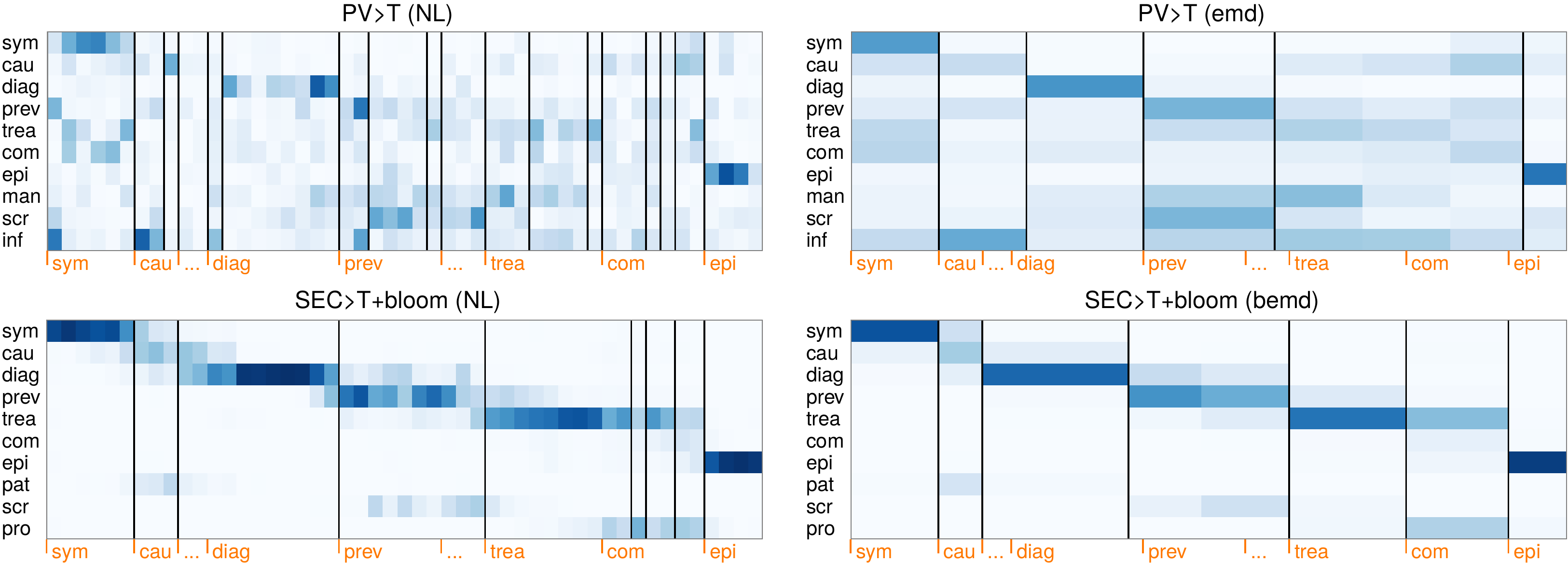}\\
  \caption{Heatmaps of predicted topic labels $\mathbf{\hat{y}_k}$ for document \example{Trichomoniasis} from PV and \textsc{Sector} models with newline and embedding segmentation. Shading denotes probability for 10 out of 27 selected topic classes on Y axis, with sentences from left to right. Segmentation is shown as black lines, X axis shows expected gold labels. Note that segments with same class assignments are merged in both predictions and gold standard (`\ldots').
  }\label{fig:heatmaps}
\end{figure*}

Table \ref{tab:results_classes} shows the evaluation results of the \textsc{WikiSection}-topics single-label classification task, Table \ref{tab:results_headings} contains the corresponding numbers for multi-label classification. Table \ref{tab:results_segments} shows results for topic segmentation across different datasets.

\paragraph{\textsc{Sector} outperforms existing classifiers.}

With our given segmentation baseline (NL), the best sentence classification model CNN achieves 52.1\% F1 averaged over all datasets. \textsc{Sector} improves this score significantly by 12.4 points. Furthermore, in the setting with plain text input, \textsc{Sector} improves the CNN score by 18.8 points using identical baseline segmentation. Our model finally reaches an average of 61.8\% F1 on the classification task using sentence embeddings and bidirectional segmentation. This is a total improvement of 27.8 points over the CNN model.

\paragraph{Topic embeddings improve segmentation.}

\textsc{Sector} outperforms C99 and TopicTiling significantly by 16.4 respectively 18.8 points $P_k$ on average. Compared to the maximum label baseline, our model gains 3.1 points by using the bidirectional embedding deviation and 1.0 points using sentence embeddings. Overall, \textsc{Sector} misses only 4.2 points $P_k$ and 2.6 points F1 compared to the experiments with prior newline segmentation.
The third experiments reveals that our segmentation method in isolation almost reaches state-of-the-art on existing datasets and beats the unsupervised baselines, but lacks performance on cross-dataset evaluation.

\paragraph{Bloom filters on par with word embeddings.}

Bloom filter encoding achieves high scores among all datasets and outperforms our bag-of-words baseline, possibly because of larger training batch sizes and reduced model parameters. 
Surprisingly, word embeddings did not improve the model significantly. On average, German models gained 0.7 points F1 while English models declined by 0.4 points compared to bloom filters. However, model training and inference using pre-trained embeddings is faster by an average factor of 3.2.

\paragraph{Topic embeddings perform well on noisy data.}

In the multi-label setting with unprocessed Wikipedia headings, classification precision of \textsc{Sector} reaches up to 72.3\% P@1 for 2.8k labels. This score is in average 9.5 points lower compared to the models trained on the small number of 25–30 normalized labels. Furthermore, segmentation performance is only missing 3.8 points $P_k$ compared to the topics task. Ranking loss could not improve our models significantly, but achieved better segmentation scores on the headings task. Finally, the cross-domain English \emph{fullwiki} model performs only on baseline level for segmentation, but still achieves better classification performance than CNN on the English cities dataset.

\subsection{Discussion and Model Insights}

Figure \ref{fig:heatmaps} shows classification and segmentation of our \textsc{Sector} model compared to the PV baseline. 

\paragraph{SECTOR captures latent topics from context.}
We clearly see from NL predictions (left side of Figure \ref{fig:heatmaps}) that \textsc{Sector} produces coherent results with sentence granularity, with topics emerging and disappearing over the course of a document. In contrast, PV predictions are scattered across the document. Both models successfully classify first (\example{symptoms}) and last sections (\example{epidemiology}). However, only \textsc{Sector} can capture \example{diagnosis}, \example{prevention} and \example{treatment}. Furthermore, we observe additional \example{screening} predictions in the center of the document. This section is actually labeled \example{"Prevention | Screening"} in the source document, which explains this overlap.

Furthermore, we observe low confidence in the second section labeled \example{cause}. Our multi-class model predicts for this section $\{$\example{diagnosis}, \example{cause}, \example{genetics}$\}$. The ground truth heading for this section is \example{"Causes | Genetic sequence"}, but even for a human reader this assignment is not clear. This shows that the multi-label approach fills an important gap and can even serve as an indicator for low-quality article structure.

Finally, both models fail to segment the \example{complication} section near the end, because it consists of an enumeration. The embedding deviation segmentation strategy (right side of Figure \ref{fig:heatmaps}) completely solves this issue for both models. Our \textsc{Sector} model is giving nearly perfect segmentation using the bidirectional strategy, it only misses the discussed part of \example{cause} and is off by one sentence for the start of \example{prevention}. Furthermore, averaging over sentence-level predictions reveals clearly distinguishable section class labels.
\section{Conclusions and Future Work}
\label{sec:summary}

We presented \textsc{Sector}, a novel model for coherent text segmentation and classification based on latent topics. We further contributed \textsc{WikiSection}, a collection of four large datasets in English and German for this task. Our end-to-end method builds on a neural topic embedding which is trained using Wikipedia headings to optimize a BLSTM classifier. We showed that our best performing model is based on sparse word features with bloom filter encoding and significantly improves classification precision for 25–30 topics on comprehensive documents by up to 29.5 points F1 compared to state-of-the-art sentence classifiers with baseline segmentation. We used the bidirectional deviation in our topic embedding to segment a document into coherent sections without additional training. Finally, our experiments showed that extending the task to multi-label classification of 2.8k ambiguous topic words still produces coherent results with 71.1\% average precision.

We see an exciting future application of \textsc{SECTOR} as a building block to extract and retrieve topical passages from unlabeled corpora, such as medical research articles or technical papers. One possible task is WikiPassageQA \cite{cohen2018wikipassageqa}, a benchmark to retrieve passages as answers to non-factoid questions from long articles.
\iftaclpubformat
\section*{Acknowledgements}

We would like to thank the editors and anonymous reviewers for their helpful suggestions and comments. 
Our work is funded by the German Federal Ministry of Economic Affairs and Energy (BMWi) under grant agreement 01MD16011E (Medical Allround-Care Service Solutions) and H2020 ICT-2016-1 grant agreement 732328 (FashionBrain).
\fi

\bibliography{sector_bibliography}
\bibliographystyle{acl_natbib}

\end{document}